\documentclass[twoside,UTF8]{article}
\usepackage[english]{babel}

\usepackage{PRIMEarxiv}

\usepackage[utf8]{inputenc} 
\usepackage[T1]{fontenc}    
\usepackage{hyperref}       
\usepackage{url}            
\usepackage{booktabs}       
\usepackage{amsfonts}       
\usepackage{nicefrac}       
\usepackage{microtype}      
\usepackage{lipsum}
\usepackage{fancyhdr}       

\usepackage{graphicx}       
\graphicspath{{media/}}     

\title{

When Graph Data Meets Multimodal: A New Paradigm for Graph Understanding and Reasoning

\thanks{This project is in progress.} 
}

\author{
  Qihang Ai, Jianwu Zhou, Haiyun Jiang\thanks{\hspace{1mm} Haiyun Jiang is the corresponding author.}, Lemao Liu, Shuming Shi \\
  Tencent AI Lab \\
\texttt{\{qihang-ai,1120211477\}@bit.edu.cn,
\{haiyunjiang,redmondliu,shumingshi\}@tencent.com} \\
}

\begin{document}
\maketitle

\begin{abstract}
Graph data is ubiquitous in the physical world, and it has always been a challenge to efficiently model graph structures using a unified paradigm for the understanding and reasoning on various graphs. 
Moreover, in the era of large language models, integrating complex graph information into text sequences has become exceptionally difficult, which hinders the ability to interact with graph data through natural language instructions.
The paper presents a new paradigm for understanding and reasoning about graph data by integrating image encoding and multimodal technologies. 
This approach enables the comprehension of graph data through an instruction-response format, utilizing GPT-4V's advanced capabilities. The study evaluates this paradigm on various graph types, highlighting the model's strengths and weaknesses, particularly in Chinese OCR performance and complex reasoning tasks. 
The findings suggest new direction for enhancing graph data processing and natural language interaction.
\end{abstract}


\section{Introduction}


Graph data is ubiquitous in the real world, illustrated by a range of examples from social networks\cite{wu2020graph} to drug discovery\cite{fout2017protein}. 
It transforms complex information into clear structural formats, which are crucial for understanding and effective communication in areas such as recommendation systems and spatiotemporal forecasting. 
The understanding and reasoning on graph structures are vital for various downstream tasks \cite{lin2023zero,chen2022learning,wang2022entity}  and essential for the 
progress of Artificial General Intelligence.

Graph data is commonly modelled using Graph Networks\cite{scarselli2008graph, bruna2013spectral,leskovec2006sampling,wang2015local,grover2016node2vec,sperduti1997supervised,hamilton2017inductive,goyal2018graph,gonzalez2014graphx,low2012distributed}, with a wealth of seminal literature supporting this approach. 
However, each type of graph, such as knowledge graphs or mind maps, has unique structures and characteristics that present challenges to a unified approach. 
Traditional graph processing methods, such as those based on matrices or classical algorithms, often focus on easy graph data and struggle to address the diversity and complexity of different graph types\cite{bronstein2017geometric}. 
In recent years, graph Neural Networks\cite{li2015gated,dai2018learning,battaglia2018relational,fan2019graph,zhang2018link} provide a more flexible and effective means of dealing with this diversity and complexity. 
By propagating and aggregating node information within the graph structure, GNNs capture complex graph features. 
Variants such as Graph Convolutional Networks (GCNs)\cite{defferrard2016convolutional,li2018deeper,chen2018fastgcn} and Graph Attention Networks (GATs)\cite{velivckovic2017graph,wang2019heterogeneous} manage relationships and information aggregation between nodes in different ways, allowing GNNs to adapt to different graph types\cite{kipf2016semi}.
Despite the ability of GNNs to handle different types of graph data, they face challenges such as scalability and limitations in broader graph data processing scenarios. 
A key issue for each graph type is the need to train a new GNN model for that specific graph, highlighting a limitation in their application.

Furthermore, integrating graph data and text data poses a significant challenge due to the inherent structural information in graph data and the sequential nature of large language models\cite{chai2023graphllm,chen2022learning}. 
Merging the encoding features of graph neural networks with language models is a formidable task. Alternatively, modifying the architecture of large language models to better incorporate features from graph neural networks is a possible solution. However, this approach presents the significant challenge of retraining the models from scratch. Such a process is not only complex, but also resource intensive, with costs that can easily exceed several hundred thousand dollars.

Therefore, we propose a new paradigm to address the two major challenges identified: (1) the difficulty in uniformly modeling various rich graph data, and (2) the complexity of integrating graph data information into language models for understanding and reasoning about graph data. 
We aim to process graph data using image encoding techniques and then leverage multimodal technology to merge graph and text data. 
This integration enables access to and comprehension of graph data through a instruction-response format. 
We aim to investigate the effectiveness of GPT-4V in this context, given its exceptional capabilities, particularly its impressive performance with everyday images\cite{yang2023dawn}. 
The innovative application of GPT-4V in this field may provide new insights and advancements in data processing and language model integration.

\begin{figure}[ht]
    \centering
    \begin{minipage}[b]{0.45\textwidth}
        \includegraphics[width=\textwidth]{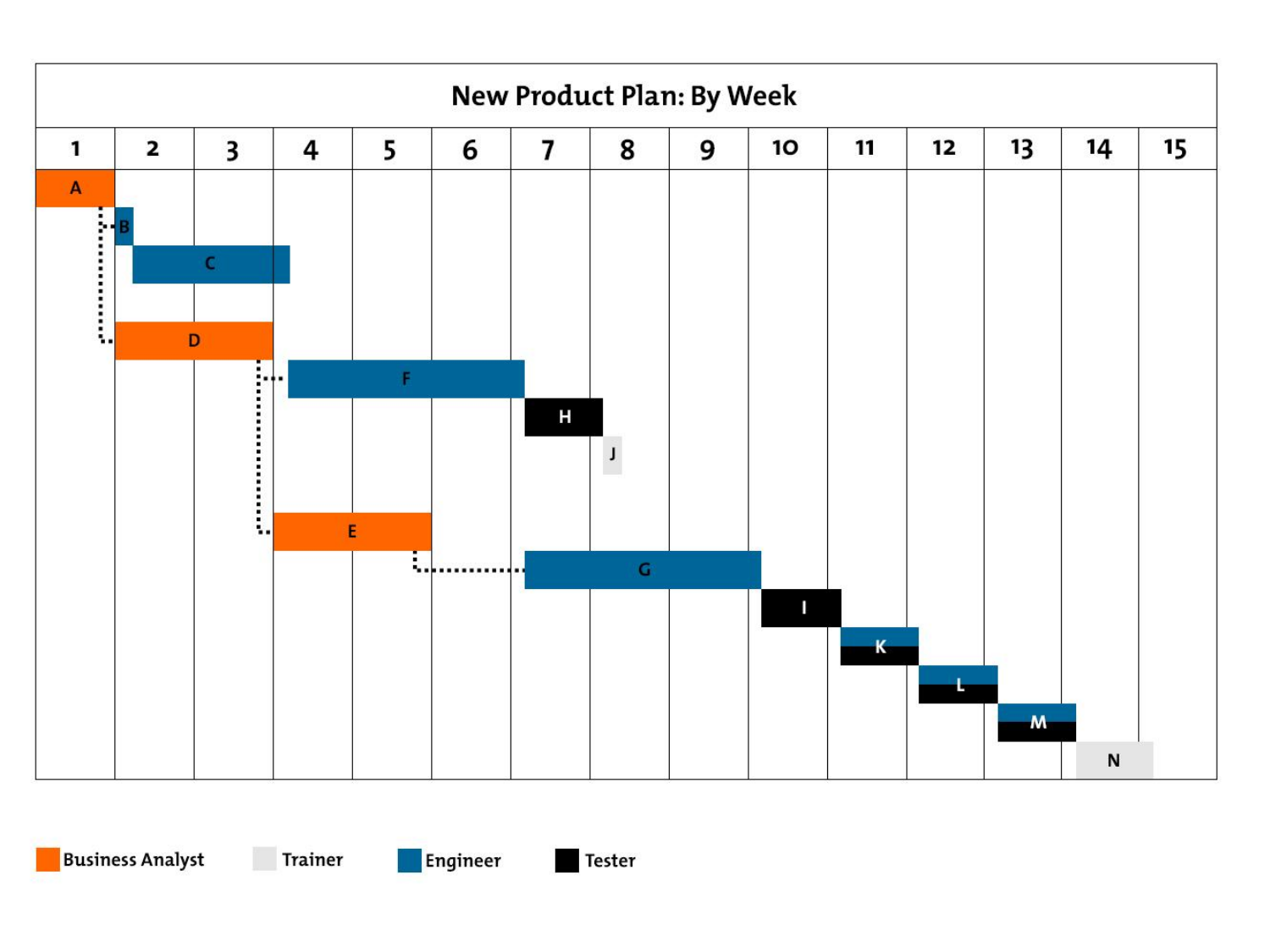}
        \caption{An example of Gantt chart.}
        \label{fig:gantt_intro}
    \end{minipage}
    \hfill
    \begin{minipage}[b]{0.45\textwidth}
        \includegraphics[width=\textwidth]{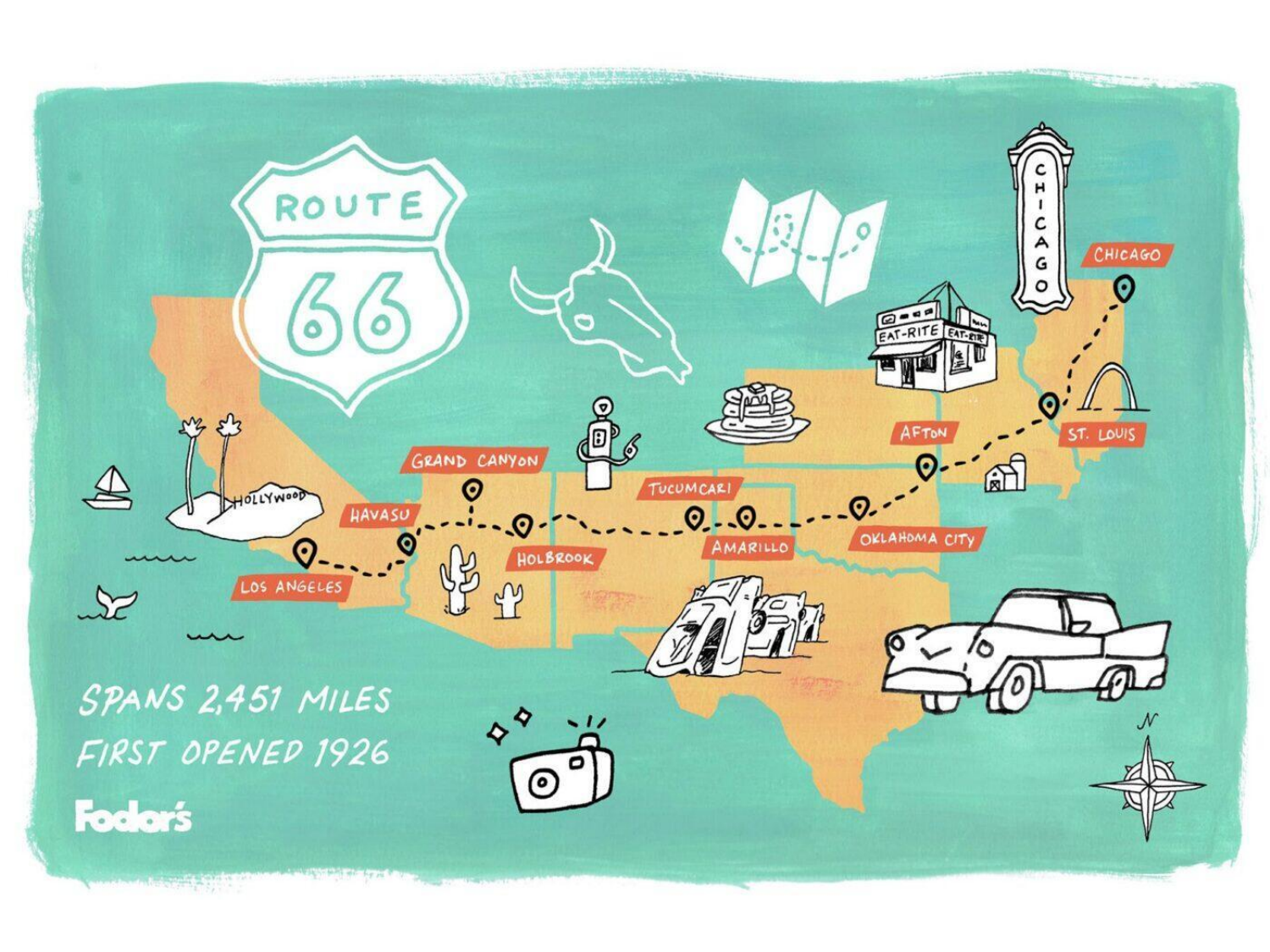}
        \caption{An example of route map.}
        \label{fig:routemap_intro}
    \end{minipage}
\end{figure}

In this initial exploration, the paper considers several typical representative graphs: knowledge graphs, flowcharts, mind maps, route maps, and Gantt charts. 
A Gantt chart, as depicted in Figure \ref{fig:gantt_intro}, is a visual tool for project management that displays the timeline of tasks and their durations. It is often color-coded by roles or stages and is an efficient way to track responsibilities. On the other hand, a route map, such as the one depicted in Figure \ref{fig:routemap_intro}, is a simplified diagram that outlines a path from one point to another, emphasizing important landmarks and destinations. It is useful for navigation and travel planning.

\begin{figure}[ht]
    \centering
    \begin{minipage}[b]{0.45\textwidth}
        \includegraphics[width=\textwidth]{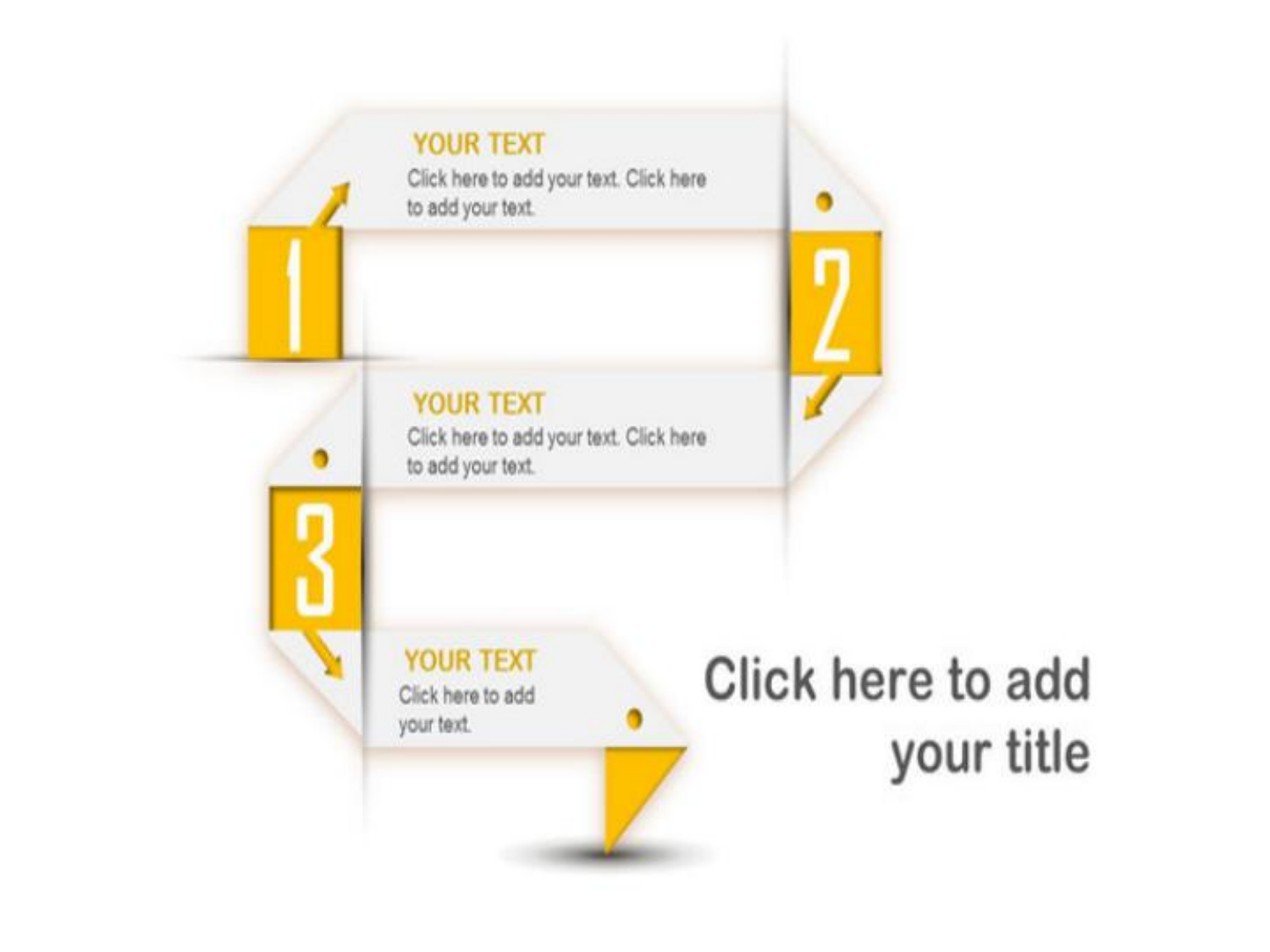}
        \caption{An example of flow chart.}
        \label{fig:image1}
    \end{minipage}
    \hfill
    \begin{minipage}[b]{0.45\textwidth}
        \includegraphics[width=\textwidth]{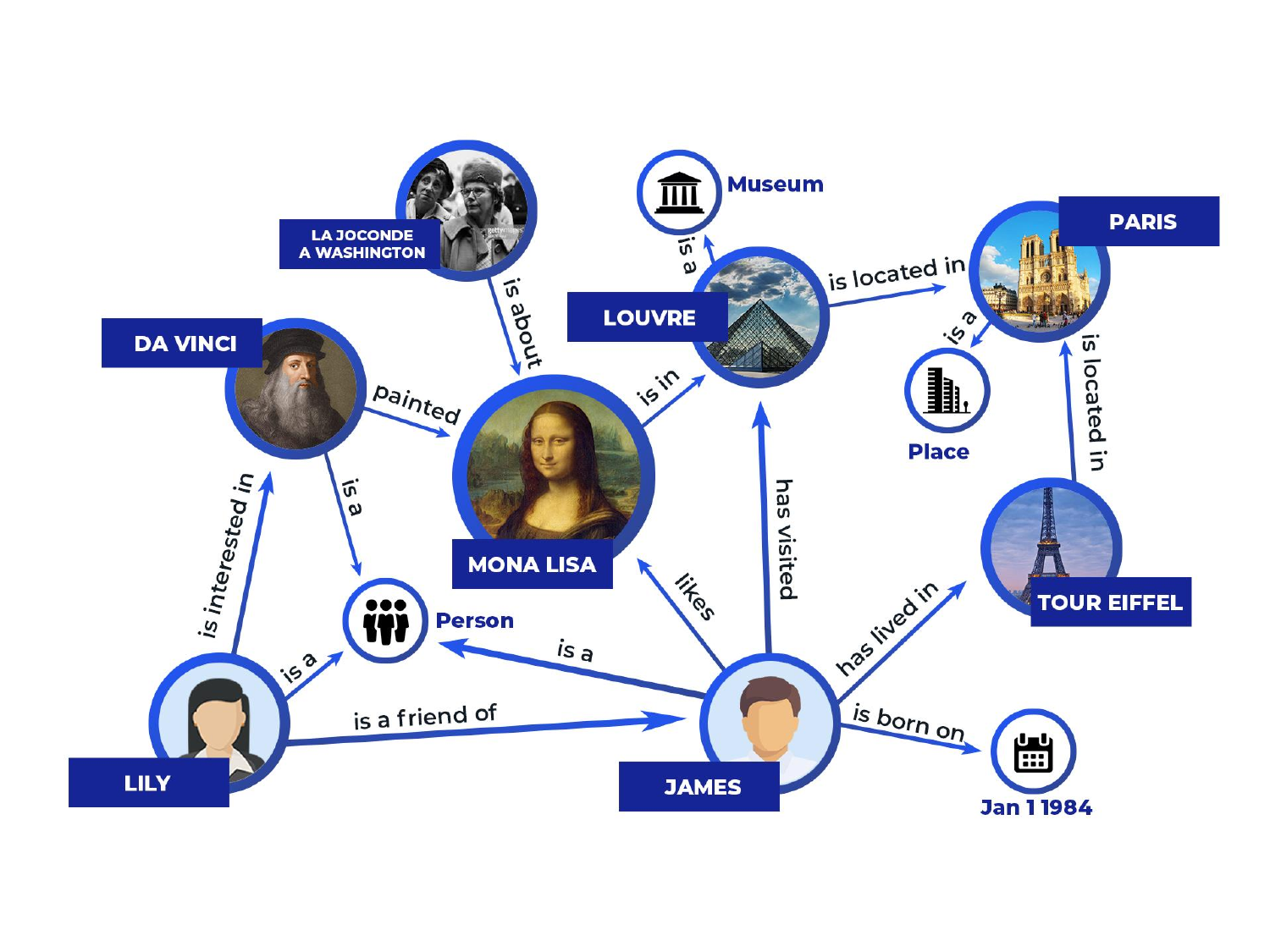}
        \caption{An example of knowledge graph.}
        \label{fig:image3}
    \end{minipage}
\end{figure}

\begin{figure}
    \centering
    \includegraphics[width=0.5\textwidth]{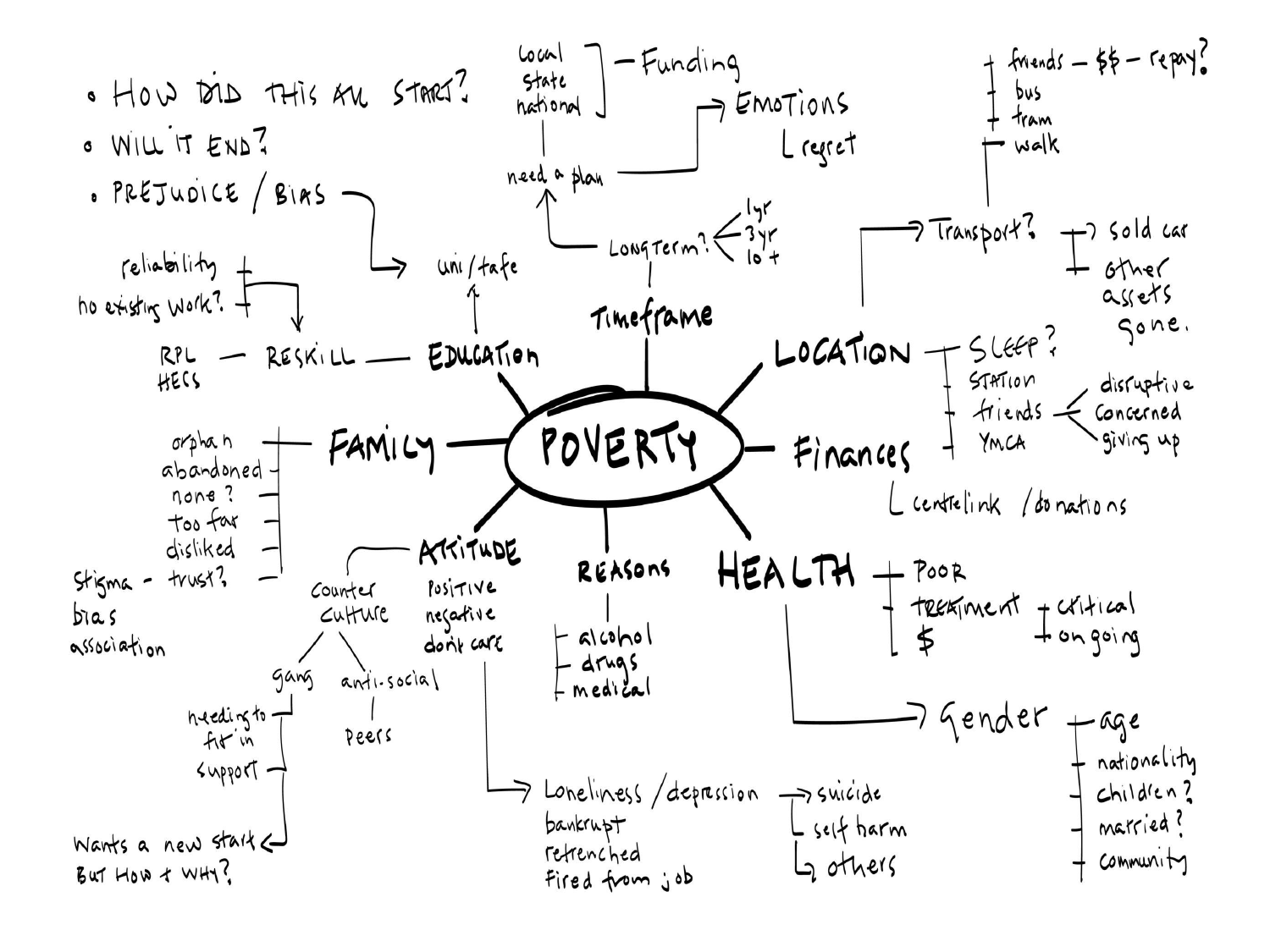}
    \caption{An example of mindmap.}
    \label{fig:gantt_intro}
\end{figure}



The evaluation process involved acquiring images from the internet, followed by a manual selection to filter out those that are clear and relevant to the topic. 
The evaluation of these images was conducted through two main approaches: command generation and command adherence. 
Command generation focuses on assessing the effectiveness of GPT-4V in generating relevant instructions. 
Command adherence, on the other hand, comprehensively explores GPT-4V's ability to follow instructions, considering factors such as the complexity of the commands, image noise, and information density. 
This approach provides a thorough examination of GPT-4V's capabilities in understanding and interacting with various types of graph data.

In conclusion, our evaluation of various real-world images reveals that GPT-4V possesses a certain degree of capability in understanding and reasoning about images. However, there are still some issues present. To summarize, these problems include a lack of proficiency in Chinese OCR and subpar performance with specific types of images, such as Gantt charts and route maps. This paper leverages the robust processing power and multimodal functionalities of GPT-4V to enhance the understanding and reasoning abilities concerning various graph data. This approach potentially paves the way for new possibilities in graph data processing, offering a more effective and flexible solution in the face of the diversity and complexity of graph data.








\section{Evaluation on GPT-4}

\subsection{Evaluation Setting}

\subsubsection{Graph Image Acquisition}
The images for our dataset were sourced from search engines. 
We conducted keyword searches in Chinese for various types such as 'knowledge graph', 'route map', 'flowchart', 'gantt chart', and 'mind map' to initially obtain images. 
Subsequently, representative images from each category were selected, and more images were retrieved using the search engine's reverse image search feature. 
After filtering out unclear images and irrelevant images which doesn't belong to any graph type, our final dataset comprised 156 knowledge graphs, 242 route maps, 157 flowcharts, 40 Gantt charts, and 31 mind maps. The proportion of Chinese to English annotated images was approximately 3:1.  

In this study, our evaluation approach centers around a multimodal data format. 
Specifically, the process involves presenting images to GPT-4 and prompting it to generate a series of instructions, followed by corresponding responses. 
The format of GPT-4's responses is denoted as \textit{\{Q\textsubscript{1}, A\textsubscript{1}, Q\textsubscript{2}, A\textsubscript{2}, ...\}}, where \textit{Q\textsubscript{i}} represents a question and \textit{A\textsubscript{i}} represents the corresponding answer.


\subsubsection{What Ability Are We Focus On?}

\paragraph{Assessment of Instruction Validity.}
The initial phase of our study involved a manual evaluation of GPT-4's capability to generate instructions in response to images. 
This crucial step was centered around assessing the relevance of the instructions to the images.
We determined the validity of each instruction based on its contextual alignment with the corresponding image, ensuring that the generated instructions were pertinent and appropriate to the visual content.

\paragraph{Evaluation of Response Validity.}
After identifying instructions that demonstrated relevance to their respective images (termed 'validInstructions'), these were selected for a more detailed analysis. 
We evaluated the correctness of the candidate answers generated by GPT-4 for these instructions from multiple perspectives: 

\begin{itemize}
    \item Simple and medium instruction following ability: Examining GPT-4's ability to accurately follow easy and intermediate-level instructions which typically addressed individual elements.
    \item Complex instruction following ability: Gauging the model's proficiency with more complex instructions requiring higher-order comprehension and understanding.
    \item Multi-hop reasoning ability: Testing the model's capability for reasoning that involves multiple inferential steps.
    \item Robustness to noise: Assessing the model's resilience in the face of potential ambiguities in the data.
    \item Performance with different information densities: Various diagrams contain varying content and details, including differences in text density. Hence, it is important to assess the ability of GPT-4V to recognise and comprehend pictures with different densities.
    \item Ablation study on different types of graphs: Investigating the model's performance  on each type of graph data, to be specific, knowledge graphs, route maps, flowcharts, gantt charts, and mind maps.
\end{itemize}

Complex instruction following tasks require understanding intricate commands and may result in simple answers. In contrast, "multi-hop reasoning" involves linking several logical steps, usually leading to more elaborate responses.

The evaluation metric focused on the validity of instructions generated by GPT-4V and the accuracy of its responses to the valid instructions. 
This was determined through manual judgment by human evaluators. 
The accuracy across the various aspects mentioned above was calculated to provide a quantitative measure of GPT-4's performance. 

\subsection{Prompt Design}
The design of our prompt, inspired by the methodology in \cite{yang2023dawn}, emphasizes two key aspects: expert identification and ensuring correctness. To illustrate, consider the prompt tailored for a knowledge graph:

\textit{"You are an expert in the field of knowledge graph. You must ensure that the image is correctly identified, and that the questions raised and answers are correct. Please ask and answer some questions based on this image."}

Initially, this prompt led GPT-4V to produce questions that were mostly short and straightforward, often lacking in complex reasoning or depth, such as multi-hop inference. 

To refine the prompt and enhance its effectiveness, we introduced additional specifications:

\textit{"You are an expert in the field of knowledge graph. You must ensure that the image is correctly identified, and that the questions raised and answers are correct. Please ask and answer some questions based on this image. The questions should preferably be based on specific nodes/edges of the image and should range from simple to complex."}

Based on the meticulously designed English prompt, for images with Chinese content, we adapted the prompt accordingly. 


\begin{figure}
    \centering
    \includegraphics[width=1.0\textwidth]{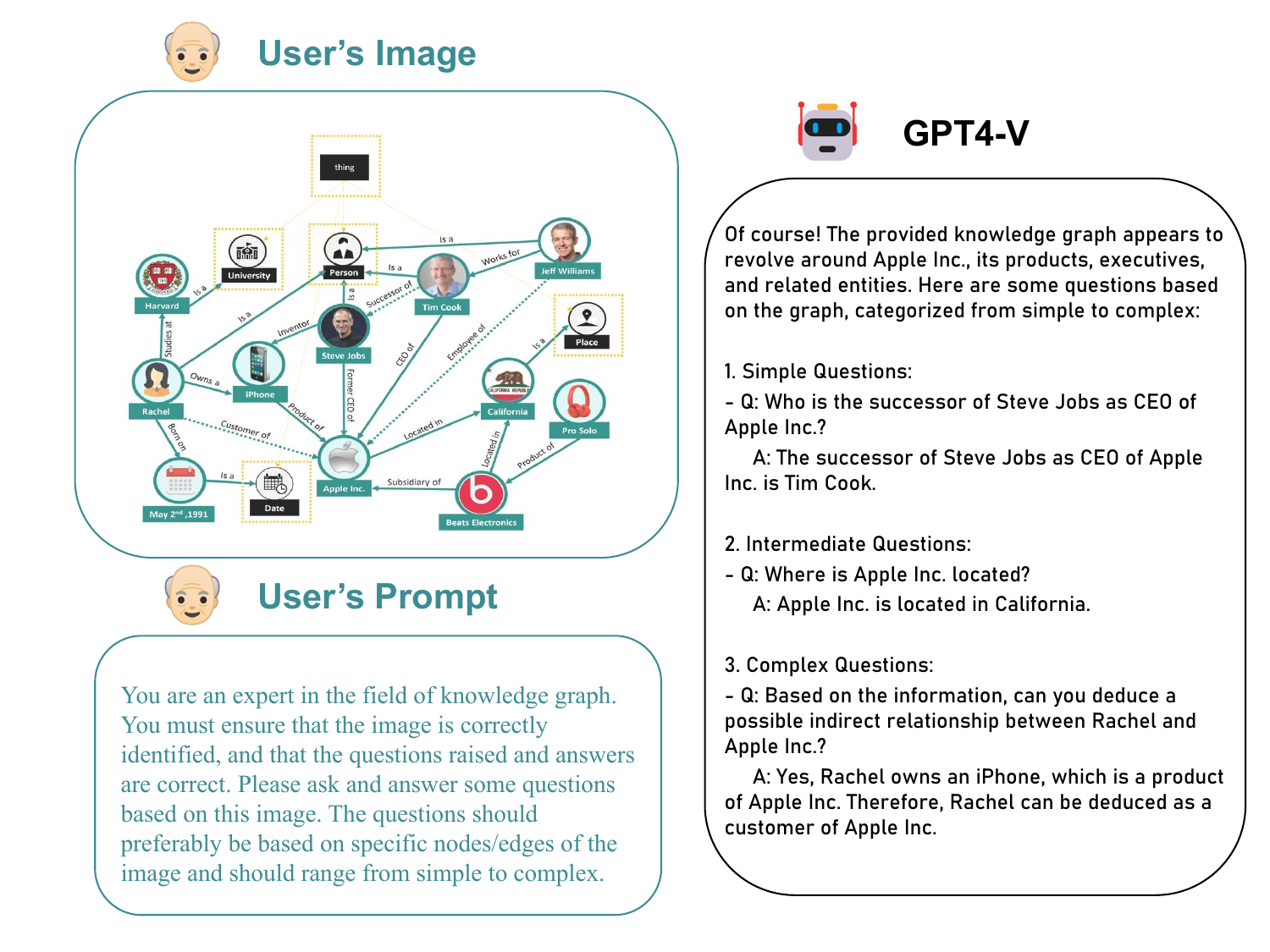}
    \caption{The instructions and answers followed by the user prompt for a given image in English.}
    \label{fig:eng}
\end{figure}

\begin{figure}
    \centering
    \includegraphics[width=1.0\textwidth]{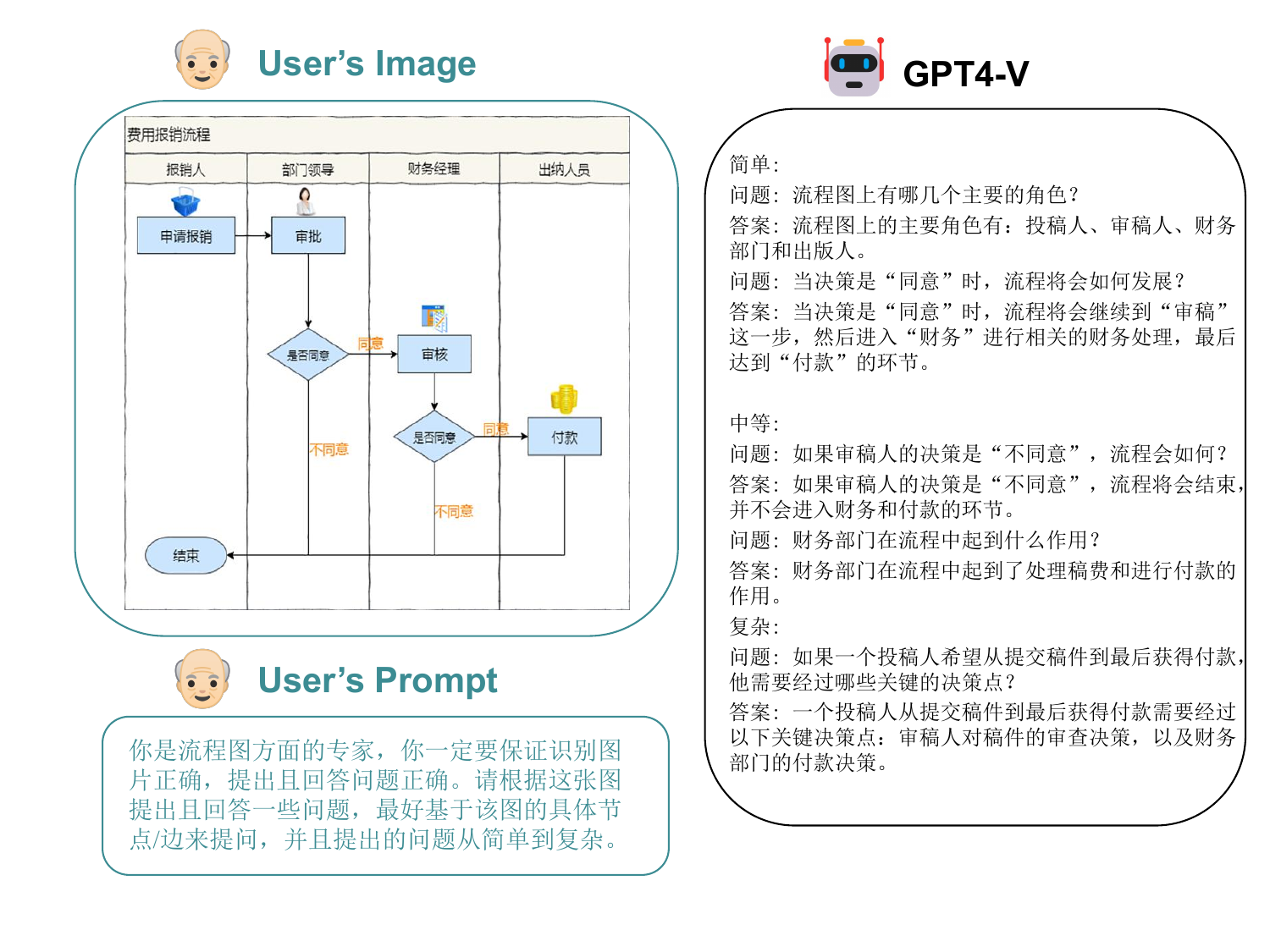}
    \caption{The instructions and answers followed by the user prompt for a given image in Chinese.}
    \label{fig:chi}
\end{figure}

For a variety of graph types, such as route maps, flowcharts, Gantt charts, and mind maps, we customized the prompts by substituting 'knowledge graph' in the original prompt with the specific type of graph being addressed. 
This adaptation was crucial to cater to the unique characteristics and requirements of each graph type.

The refined prompts, incorporating instructions to focus on specific nodes/edges and to vary the complexity of questions, significantly improved the quality of the generated questions. 
This change was instrumental in overcoming the initial limitation of generating overly simple and brief instructions. 
By directing the focus to precise elements within each graph and encouraging a range of complexities, we ensured that the questions were not only accurate but also varied in depth, moving from simple factual inquiries to more complex, inferential ones.

Figure \ref{fig:eng} presents an example of a knowledge graph with English annotations, where the instructions and responses were generated by GPT-4V following the specified English prompt. 
Similarly, Figure \ref{fig:chi} illustrates a flowchart with Chinese annotations, with instructions and responses produced by GPT-4V in accordance with the adapted Chinese prompt.





\subsection{Instruction Generation Ability}

In this subsection, we evaluate GPT-4's ability to generate instructions in response to visual stimuli. 
Our dataset comprises 50 images, evenly balanced between Chinese and English annotations, with 25 in Chinese and 25 in English. 
We tasked GPT-4 with generating instructions for each image. 
Guided by our prompt, GPT-4V autonomously categorized the complexity of these instructions within its responses, labeling them as simple, medium, or complex.
Typically, 1-3 questions were generated for each level of complexity, resulting in a total of 3-9 instructions per image.

\subsubsection{Relevance Assessment}
The relevance of the generated instructions to the images was evaluated. 
The relevance of instructions generated by GPT-4 was critically analyzed. Despite all images being valid, the generated instructions will be classified as invalid based on two criteria.

\begin{itemize}
    \item Instructions completely unrelated to the image's theme.
    \item Instructions related to the image's theme but involving nodes or edges not present in the image.
\end{itemize}


Given an image, an irrelevant question will be labeled as an invalid case. The distribution of valid and invalid instructions is detailed in Table \ref{tab:validity_assessment_instruction}.

\begin{table}[h]
\centering
\begin{tabular}{lccc}
\toprule
\textbf{Language} & \textbf{Valid Instructions} & \textbf{Invalid Instructions} \\ 
\midrule
English           & 146 & 4 \\ 
Chinese           & 45 & 102 \\ 
\bottomrule
\end{tabular}
\caption{Validity assessment of instructions generated by GPT-4V.}
\label{tab:validity_assessment_instruction}
\end{table}

The error rate was particularly high in instructions generated for images annotated in Chinese, while it's quite low in English images. 
At first, we suspected that there might be issues with GPT's Chinese-English translation in regards to image recognition. However, we couldn't explain those instructions that were completely unrelated to the theme of the image. Therefore, we speculate that there might be problems with the Chinese OCR of GPT-4V.
To validate this, we conducted further tests on the OCR capabilities of GPT-4 with Chinese images.

\subsubsection{OCR Performance Test}
To further investigate GPT-4's OCR capabilities, a test was conducted on images annotated in Chinese using the following prompt:
"Please act as an expert in Chinese and use GPT-4's image recognition ability to identify all the text in the image, ensuring the correct answer". 
The OCR result is presented in Table \ref{tab:ocr_accuracy}.

\begin{table}[h]
\centering
\begin{tabular}{cc}
\toprule
\textbf{Total Recognitions} & \textbf{Correct Recognitions} \\ 
\midrule
114                         & 34                           \\ 
\bottomrule
\end{tabular}
\caption{Evaluation results of GPT-4V OCR capability for chinese characters.The unit of measurement for the evaluation is based on the number of recognized nodes or edges.}
\label{tab:ocr_accuracy}
\end{table}

We extracted one image each from five categories: knowledge graphs, route maps, flowcharts, Gantt charts, and mind maps, all annotated in Chinese. The OCR accuracy was calculated by comparing the number of correct recognitions of nodes and edges as identified by GPT-4 against the total number of recognitions. 
It was found to be approximately 29.8\%. 
Our analysis indicates that some of the errors in GPT-4's responses included misidentifications that were semantically or structurally similar to the correct characters.
Additionally, GPT-4's OCR results for Chinese text occasionally produced inexplicable errors, such as interpreting "Asia-Pacific Economic Cooperation" as "Liubotaitaitaitai" (a nonsensical string of characters).

\subsection{Simple and Medium Instruction Following Ability}
\label{sec:simple_medium}


In our initial analysis, we found little difference between simple and medium complexity questions generated by GPT-4V. Therefore, we combined them for analysis in this section.
We randomly sampled ten images from the data set which together contained 55 valid instructions categorized as either simple or intermediate. 
The assessment of GPT-4V's responses to these instructions are presented in Table \ref{tab:instruction_accuracy}.

\begin{table}[h]
\centering
\begin{tabular}{lccc}
\toprule
\textbf{Instruction Complexity} & \textbf{Correct Responses} & \textbf{Total Instructions} & \textbf{Accuracy Rate} \\
\midrule
Simple                          & 21                          & 27                          & 77.8\%                 \\ 
Medium                          & 22                          & 28                          & 78.6\%                 \\ 
\bottomrule
\end{tabular}
\caption{Comparative results of response accuracy for simple and medium complexity instructions.}
\label{tab:instruction_accuracy}
\end{table}

The result confirms the validity of our decision to merge simple and medium complexity instructions for collective analysis, as GPT-4V's response accuracy rates are similar for both categories.
Despite the overall high accuracy, we observed a lower performance on Gantt charts. 
GPT-4V struggled with correctly aligning the bar representations of time and tasks, leading to errors in determining the start times of tasks and their overlaps. 
This issue is further discussed in detail in Section \ref{sec:ablation_study}, where Figure \ref{fig:gantt} provides a visual representation of the challenges faced by GPT-4V when processing Gantt charts.
Other errors included misrecognition of node names, which led to incorrect responses. 
This node misrecognition manifested in two ways: first, the identification of node names was incorrect, and second, the understanding of the node's meaning was inaccurate. 
Additionally, errors in the number of subnodes, such as omitting or spuriously adding subnodes, were noted in the responses.

\subsection{Complex Instruction Following Ability}


We randomly sampled ten images from our data set and focus on GPT-4's responses to complex instructions. 
In addressing complex questions, GPT-4V exhibited errors in node value recognition. Moreover, as such questions often involve transitions across multiple nodes and relationships, GPT-4V also demonstrated inaccuracies in edge value relationship judgments, leading to logical misjudgments between nodes.

\begin{table}[h]
\centering
\begin{tabular}{lccc}
\toprule
\textbf{Instruction Complexity} & \textbf{Correct Responses} & \textbf{Total Instructions} & \textbf{Accuracy Rate} \\ 
\midrule
Complex                         & 18                          & 25                          & 72\%                   \\ 
\bottomrule
\end{tabular}
\caption{Results of response accuracy for complex instructions. The accuracy rate is a little lower than that of simple and medium instructions but still remains at a satisfactory level.  }
\label{tab:complex_instruction_accuracy}
\end{table}

\begin{figure}
    \centering
    \includegraphics[width=1.0\textwidth]{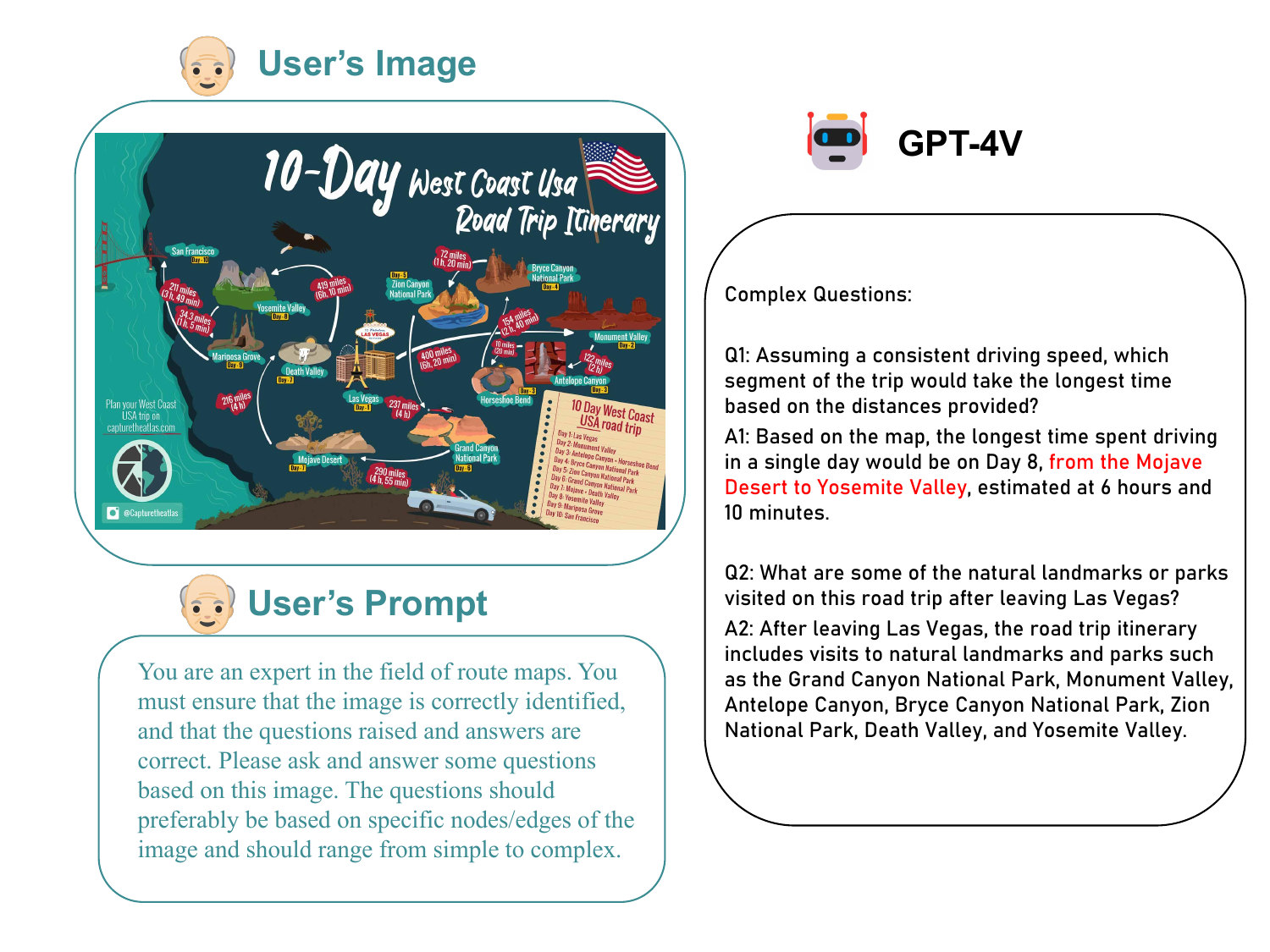}
    \caption{Complex Instruction Following for a route map. The wrong statements in responses generated by GPT-4V are marked in red.}
    \label{fig:routemap_complex}
\end{figure}


In the routemap shown in Figure \ref{fig:routemap_complex}, GPT-4V incorrectly placed the starting point of day 7 at the starting point of day 8. 
In relationship understanding, it was observed that GPT-4V tends to confuse the directions of directed edges during complex reasoning processes, resulting in erroneous relationship judgments.

\subsection{Multi-hop Reasoning Ability}




Data was gathered by randomly selecting ten images containing multi-hop reasoning questions, resulting in the evaluation of nineteen instructions. 
The results\ref{tab:multi_hop_reasoning} showed an accuracy rate of 63.2\% (12 out of 19).

\begin{table}[h]
\centering
\begin{tabular}{ccc}
\toprule
\textbf{Total Instructions} & \textbf{Correct Instructions} & \textbf{Accuracy Rate} \\ 
\midrule
19                          & 12                            & 63.2\%                 \\ 
\bottomrule
\end{tabular}
\caption{Results of response accuracy for multi-hop reasoning instructions. GPT-4V's multi-hop inference has an accuracy of 63.2\%, which is nearly 10 percentage points lower even than the accuracy followed by complex instructions. }
\label{tab:multi_hop_reasoning}
\end{table}

In analyzing GPT-4V's reasoning capabilities, we concentrated on two main areas:

\begin{itemize}
    \item Pattern Recognition Errors: GPT-4V might fail to accurately identify specific patterns or trends within the graph.
    \item Relationship Understanding Errors: Errors may occur in the AI's comprehension of logical or mathematical relationships between graphics.
\end{itemize}

\begin{figure}
    \centering
    \includegraphics[width=1.0\textwidth]{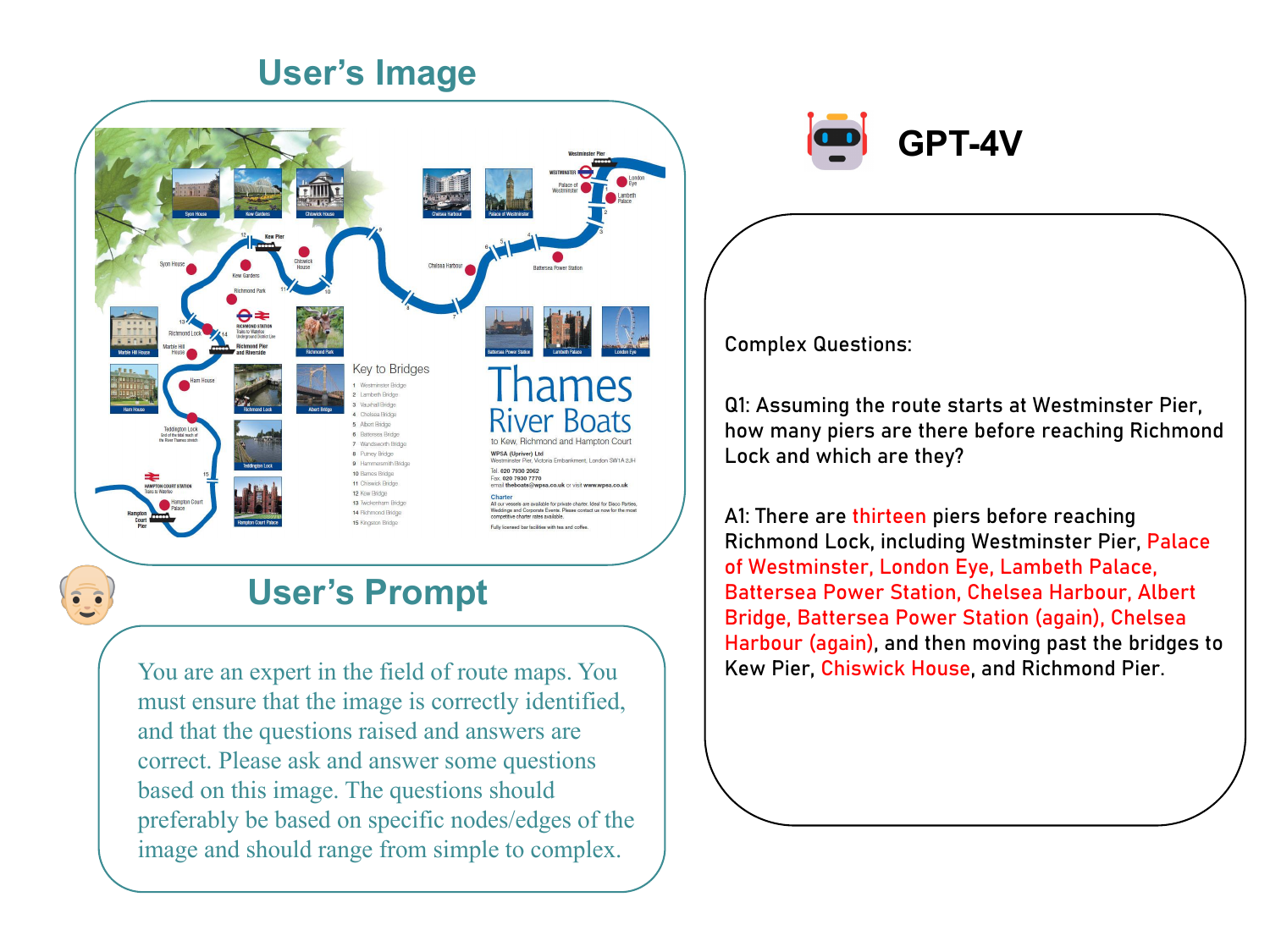}
    \caption{Multi-hop reasoning instruction following for a route map. The wrong statements in responses generated by GPT-4V are marked in red.}
    \label{fig:routemap_multihop}
\end{figure}

In route map interpretation, some errors were related to pattern recognition, where GPT-4V incorrectly identified certain landmarks or locations, such as mistaking a tourist spot for a dock in Figure \ref{fig:routemap_multihop}.
Furthermore, issues in relationship understanding were evident in GPT-4V's interpretation of numerical information. Specifically, when questions involved counting nodes, GPT-4V occasionally provided incorrect counts. 
Similarly, when mathematical expressions were required to determine the value of nodes or edges, the model's expression evaluation was found to be inaccurate. 
In the context of time-related tasks, GPT-4V faced difficulties in correctly aligning bar representations of time and tasks in Gantt charts.


\subsection{Robustness to Noise}
We randomly selected five images characterized by blurriness and noise, resulting in a total of 40 instructions for evaluation.
We initiated the evaluation by categorizing instructions into two groups: valid and invalid. An obvious increase  in the number of invalid instructions was observed for images with blurriness compared to those images without noise. 
The results are listed in Table \ref{tab:noise_validity}. 

This can be attributed to GPT-4V's limited ability to perform accurate pattern recognition. 
Consequently, it generates incorrect questions, such as misjudging the number of branching nodes or asking about non-existent nodes, rendering the instructions invalid.

\begin{table}[h]
\centering
\caption{Impact of Image Noise on Instructions by GPT-4V}
\begin{tabular}{lcccc}
\toprule
\textbf{Image Noise} & \textbf{Valid Instructions} & \textbf{Invalid Instructions} & \textbf{Valid Instruction Ratio} \\
\midrule
Yes & 30 & 10 & 75.0\% \\
No & 146 & 4 & 97.3\% \\
\bottomrule
\end{tabular}
\label{tab:noise_validity}
\end{table}

\begin{table}[h]
\centering
\caption{Impact of Image Noise on Responses by GPT-4V}
\begin{tabular}{lcccc}
\toprule
\textbf{Instruction Type} & \textbf{Correct Responses} & \textbf{Total Instructions}  & \textbf{Accuracy} \\
\midrule
Valid& 17                         & 30                           & 56.7\% \\
\bottomrule
\end{tabular}
\label{tab:noise_accuracy}
\end{table}

Then, we calculated the accuracy specifically for valid instructions.
We evaluate GPT-4V's responses for the 30 valid instructions, and the results are listed in \label{tab:noise_accuracy}. 
The decline in accuracy in answering instructions can also be linked to the model's inability to perform precise pattern recognition, leading to incorrect node values in responses. 
Image clarity serves as the foundation for GPT-4V's strong performance.

\subsection{Performance with Different Information Densities}


This section explores the performance of GPT-4V in responding to instructions for images with varying information densities. 
Images can exhibit different levels of information density, ranging from sparse to dense content. 
To investigate this, images were categorized into two groups: information-dense and information-sparse images. 
The performance of GPT-4V in responding to instructions for these two categories was then analyzed.

For this analysis, a total of 10 images were selected, representing each graph type with both an information-sparse and an information-dense image, amounting to 5 images per category. 
Each category of images was then subjected to 25 instructions, and the responses were evaluated. 
The results are summarized in the Table \ref{table:detailed_info_density_performance}.

\begin{table}[h]
\centering
\begin{tabular}{lcccc}
\toprule
\textbf{Information Density} & \textbf{Correct Responses} & \textbf{Total Instructions} & \textbf{Accuracy Rate} \\
\midrule
Dense & 16 & 25 & 64\% \\
Sparse & 22 & 25 & 88\% \\
\bottomrule
\end{tabular}
\caption{Detailed performance of GPT-4V across different information densities.}
\label{table:detailed_info_density_performance}
\end{table}

The higher error rate observed in dense images is attributed to a decrease in GPT-4V's performance in pattern recognition and relationship understanding. 
This decline in performance is influenced by two factors. 
Firstly, in the context of pattern recognition, the heightened complexity and density of nodes and edges within dense images can lead to confusion, causing GPT-4V to make incorrect associations between node and edge values, thus resulting in pattern recognition errors. 
Secondly, the high information density characteristic of dense images often stems from the diverse representation of information within a graph. Consequently, GPT-4V may encounter difficulties in fully comprehending and recognizing the intricate network of relationships present in such images, leading to errors in understanding the nuanced relationships encoded within the data.


   

\subsection{Ablation Study on Different Types of Graphs}
\label{sec:ablation_study}




\begin{figure}
    \centering
    \includegraphics[width=1.0\textwidth]{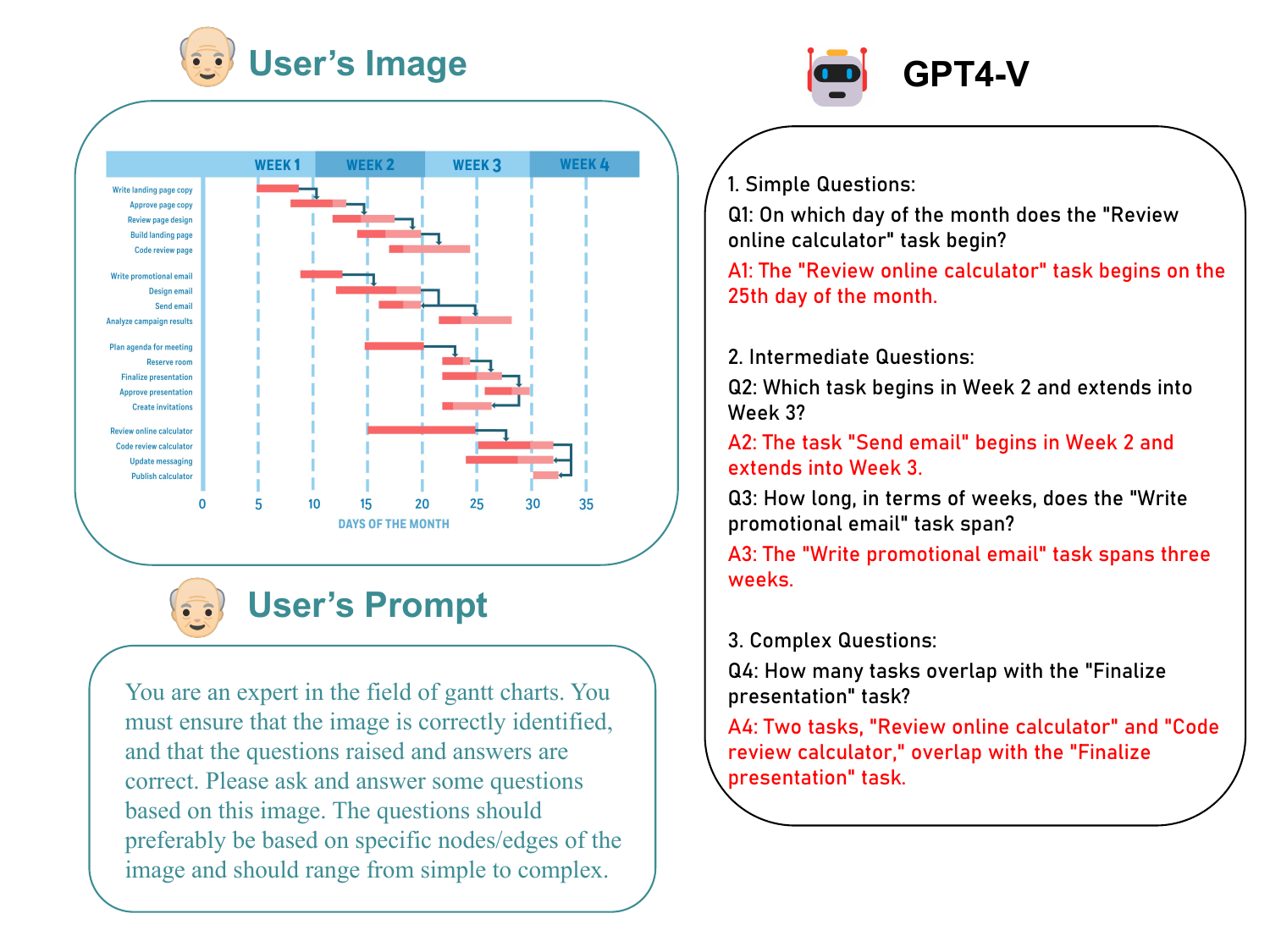}
    \caption{The instructions and answers followed by the user prompt for a gantt chart. We have selected some incorrect responses generated by GPT-4V and marked them in red.}
    \label{fig:gantt}
\end{figure}

In order to ascertain the performance of GPT-4V on each type of graph data, an ablation study was conducted. 
For each category of images, ten noise-free pictures were selected. 
Each image was subjected to valid instructions, and the correctness of the answers was manually evaluated. In total, 270 instructions were assessed, out of which 211 were answered correctly. However, GPT-4V's performance varied significantly across different types of images, as detailed below.

\begin{table}[h]
\centering
\begin{tabular}{lcccc}
\toprule
\textbf{Graph Type} & \textbf{Correct Responses} & \textbf{Total Instructions} & \textbf{Accuracy Rate} \\
\midrule
Overall & 211 & 270 & 78.1\% \\
Knowledge Graph & 77 & 81 & 95.1\% \\
Mind Map & 39 & 49 & 79.6\% \\
Route Map & 36 & 48 & 75.0\% \\
Flowchart & 44 & 60 & 73.3\% \\
Gantt Chart & 15 & 32 & 42.8\% \\
\bottomrule
\end{tabular}
\caption{Detailed accuracy rates for different types of graphs in GPT-4V.}
\label{table:detailed_accuracy_rates}
\end{table}


For Gantt Charts, it was observed that GPT-4V struggles with accurately determining overlapping times of multiple tasks and the duration of individual tasks, leading to incorrect answers. 
The model appears to be unfamiliar with aligning bar charts to timelines. 
This challenge is exemplified in Figure \ref{fig:gantt}, which showcases user prompts for a Gantt chart along with the corresponding instructions and answers. 
The figure includes a selection of incorrect responses generated by GPT-4V, highlighted in red. This visual representation (see Figure \ref{fig:gantt}) provides a clear illustration of the model's difficulty in interpreting and responding accurately to Gantt chart-related queries.

Regarding Mind Maps, errors were noted in the generation of instructions, particularly in estimating the number of sub-branches under a node, resulting in imprecise questions.
s
In the case of Route Maps, the model performed poorly in route planning. This included incorrect routes that failed to reach specified destinations and impractical route planning, involving retracing or impassable paths.

\section{Conclusion and Future Work}

    

The current study has shown a significant advancement in the field of multimodal interaction and graph data comprehension using the latest generative pre-trained transformer model, GPT-4. The investigation into graph comprehension and reasoning has revealed GPT-4V's nuanced ability to interpret, process, and engage with multimodal data, providing substantial evidence of its proficiency in executing tasks involving various graph types. Although the study has revealed potential, it has also identified challenges, particularly in areas such as Chinese OCR and specialized graph formats like Gantt charts and route maps, which have proven to be more demanding for the model. These findings confirm the capabilities of large multimodal models like the GPT-4V in handling multimodal inputs and also provide a roadmap for future work to refine these competencies. The need for improved OCR technology, particularly for non-Latin scripts, and a better understanding of complex graphical data structures is emphasized. The conclusion of this work is not just a summary of what has been accomplished, but also provides a crucial basis for future research to enhance AI's multimodal interaction and reasoning abilities.

\textbf{Data Annotation Challenges in Multimodal Datasets:} The current open-source multimodal datasets have a significant lack of graph data. To address this issue, we have initiated the annotation of a medium-scale graph dataset that includes images, instructions, and responses. The purpose of this dataset is to improve the performance of current multimodal models when processing graph data.

\textbf{Enhancing Graph Data Comprehension in Open-Source Multimodal Models:} Our analysis shows that, except for GPT-4V, the effectiveness of other open-source models, particularly in Chinese, is significantly lower. We suggest a two-pronged approach to utilize these annotated datasets to improve model performance on Chinese graph data. First, focus on processing Chinese graph data. Second, optimize the handling of less effective graph types, such as Gantt and route maps. This discussion pertains to the expected enhancements in model comprehension and inference achieved through fine-tuning.

\textbf{Advancing Chinese OCR Capabilities:} In addition to conventional instruction fine-tuning based on annotated data, it is necessary to enhance Chinese Optical Character Recognition (OCR) capabilities. This is particularly relevant in the context of multimodal models' understanding and inference of Chinese graph data. Methods for aligning Chinese text in images with its textual format counterparts are explored, and we discuss how existing Chinese OCR datasets could be used to train and refine the models.

\textbf{Understanding and reasoning about large-scale graph data:} Handling graphs with thousands of nodes and edges presents a unique challenge, as a single image often fails to adequately capture and express the full complexity of the graph. To address this, we propose to decompose such large graphs into sequential subgraphs. This approach allows us to apply multimodal methods to each subgraph. However, it requires the development of a novel capability in our models: the ability to jointly understand and infer from multiple images representing different segments of the graph. This strategy not only simplifies the complexity of large graph data, but also enables more detailed and accurate processing of the information contained within these graphs. We will explore techniques for effective image sequence analysis and how they can be integrated into existing multimodal frameworks to enhance the overall understanding and reasoning capabilities of the models.

\bibliographystyle{unsrt}  
\bibliography{references}

\end{document}